\documentclass[letterpaper, 10 pt, conference]{ieeeconf}  

\IEEEoverridecommandlockouts                              

\let\labelindent\relax
\usepackage{enumitem}
\usepackage{color,soul}

\usepackage{todonotes}
\usepackage{amsmath}
\usepackage{amssymb}
\usepackage{diagbox}
\usepackage{makecell}
\usepackage{amsmath}
\usepackage{url}
\usepackage{hyperref}
\usepackage{bm}

\def \Vbf {\mathbf{V}}
\def \Omgbf {\mathbf{\Omega}}

\newcommand{\Bframe}{\mathcal{B}}

\title{\LARGE \bf
Aerial Interaction with Tactile Sensing
}

\author{Xiaofeng Guo$^{*}$, Guanqi He$^{*}$, Mohammadreza Mousaei$^{*}$, Junyi Geng$^{\dagger}$, Guanya Shi$^{*}$, Sebastian Scherer$^{*}$
\thanks{$^{*}$Robotics Institute, Carnegie Mellon University, Pittsburgh PA 15213, USA.
$^\dagger$Department of Aerospace Engineering, Pennsylvania State University, University Park, PA, 16802, USA.
}
\thanks{Project website: \href{https://sites.google.com/view/aerial-system-gelsight}{https://sites.google.com/view/aerial-system-gelsight}}
}

\begin{document}

\maketitle
\thispagestyle{empty}
\pagestyle{empty}

\begin{abstract}

While autonomous Uncrewed Aerial Vehicles (UAVs) have grown rapidly, most applications only focus on passive visual tasks. \textit{Aerial interaction} aims to execute tasks involving physical interactions, which offers a way to assist humans in high-risk, high-altitude operations, thereby reducing cost, time, and potential hazards. The coupled dynamics between the aerial vehicle and manipulator, however, pose challenges for precision control. Previous research has typically employed either position control, which often fails to meet mission accuracy, or force control using expensive, heavy, and cumbersome force/torque sensors that also lack local semantic information. Conversely, tactile sensors, being both cost-effective and lightweight, are capable of sensing contact information including force distribution, as well as recognizing local textures. Existing work on tactile sensing mainly focuses on tabletop manipulation tasks within a quasi-static process. In this paper, we pioneer the use of vision-based tactile sensors on a fully-actuated UAV to improve the accuracy of the more dynamic aerial manipulation tasks. We introduce a pipeline utilizing tactile feedback for real-time force tracking via a hybrid motion-force controller and a method for wall texture detection during aerial interactions. Our experiments demonstrate that our system can effectively replace or complement traditional force/torque sensors, improving flight performance by approximately 16\% in position tracking error when using the fused force estimate compared to relying on a single sensor. Our tactile sensor achieves 93.4\% accuracy in real-time texture recognition and 100\% post-contact. To the best of our knowledge, this is the first work to incorporate a vision-based tactile sensor into aerial interaction tasks.
\end{abstract}



\section{INTRODUCTION}
\label{sec: intro}

Autonomous Uncrewed Aerial Vehicles (UAVs) have attracted the interest of researchers, industry, and the general public in many applications, ranging from photography and 3D mapping~\cite{zhao2021super, bonatti2020autonomous}, spraying pesticides for agriculture purpose~\cite{mogili2018review}, to package delivery~\cite{geng2020cooperative}. However, most existing studies and applications still focus on passive tasks such as visual inspection, surveillance, monitoring, etc., where the UAVs try to avoid colliding with other flying agents or obstacles such as branches or buildings without physical contact. On the other hand, many high-altitude tasks, such as changing light bulbs on tall towers, inspecting the wings or the blades of the aircraft, and bridge painting or maintenance, are still manually performed. However, climbing structures is expensive, inefficient, and hazardous to human life. \textit{Aerial interaction} intends to perform manipulation tasks with physical interaction, such as grasping, transporting, positioning, assembly, and disassembly of mechanical parts~\cite{ollero2021past, suarez2020benchmarks}, which can assist humans in such hazardous situations.

\begin{figure}
    \centering
    \includegraphics[width = \linewidth,trim={0.3cm 0.3cm 0.4cm 0cm},clip]{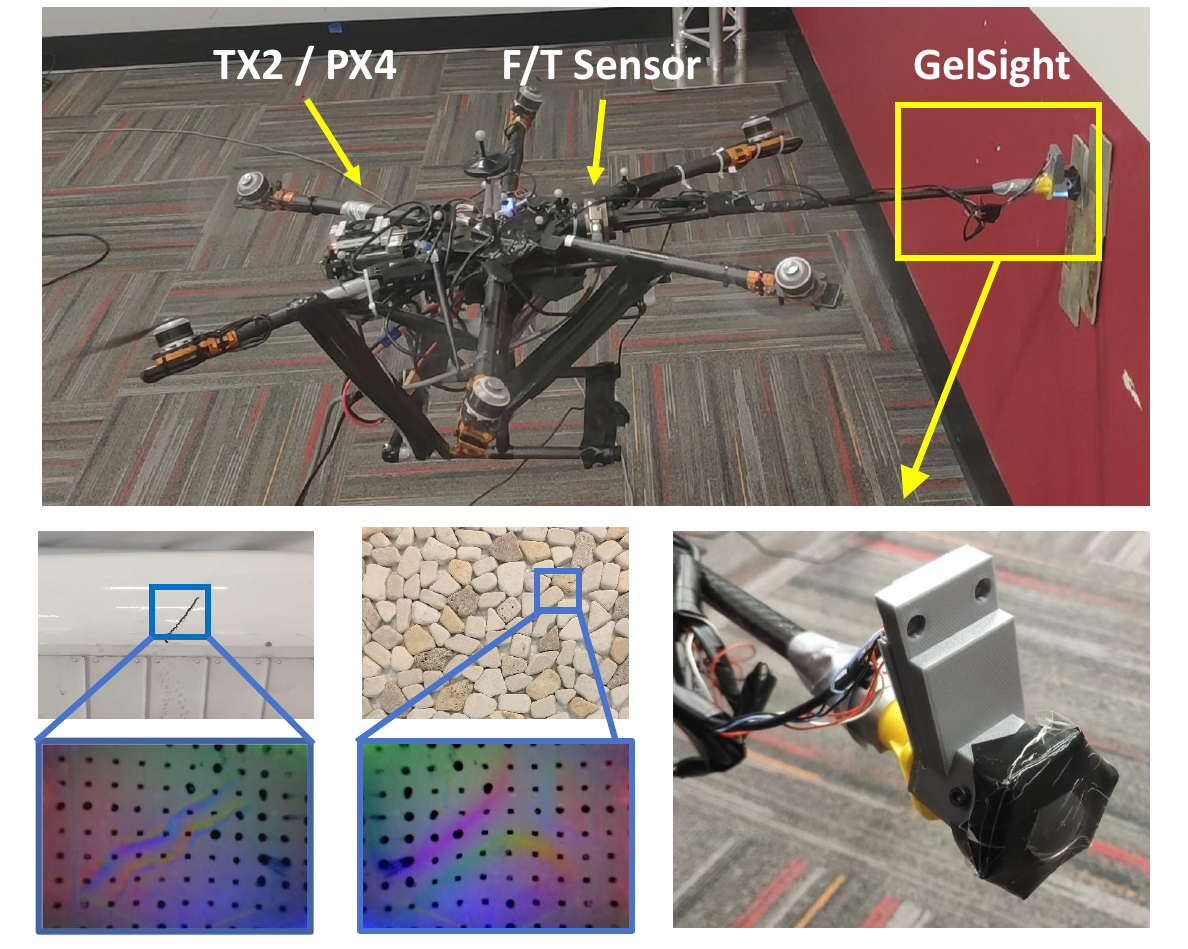}
    \caption{Overview: Integration of a vision-based tactile sensor with a fully-actuated UAV for enhanced aerial interactions. The sensor captures contact images for wall texture recognition and defect detection while also estimating contact forces, enabling the hybrid motion-force controller to accurately track target forces during the aerial interaction.}
    \label{fig: concept}
    \vspace{-2em}
\end{figure}

The key challenge in aerial manipulation lies in the interdependent behavior of the flying vehicle and the attached manipulator. In contrast to traditional robotic arms, which are often anchored to a fixed or ground-based mobile platform, aerial vehicles operate as floating bodies in 3D space. The movement of the UAV directly influences the behavior of the end-effector of the manipulator. Conversely, the interactions of the manipulator with external objects can affect the motion of the UAV. To attain the high level of precision typically required for manipulation tasks, accurate control of both the UAV and the manipulator is essential. Moreover, the inherent instability of multirotor platforms adds additional challenges under environmental disturbance, such as complex aerodynamics~\cite{o2022neural,shi2021neural}.

There are many research efforts exploring the aerial manipulation for various kinds of jobs, such as grasping~\cite{kim2013aerial,mellinger2011design}, aerial docking~\cite{choi2022automated}, or opening a door~\cite{lee2020aerial}, etc. While many of them only rely on position control~\cite{lee2020aerial, lai2022image}, tracking a reference contact force while in contact becomes an emerging trend because some applications need well-maintained force to achieve required precision, such as aerial writing~\cite{tzoumanikas2020aerial, lanegger2022aerial} or pushing a target~\cite{brunner2022energy}. Almost all of the existing work with force control rely on force torque (F/T) sensors to obtain feedback. However, most of the F/T sensors are expensive, heavy and may require a separate unit for data conversion, which adds extra complexity for mobile platform integration. People often end up mounting the F/T sensor in an indirect contact location, such as at the base of the end-effector~\cite{cuniato2022power}. This indirect sensing mechanism may lead to data delays and a lack of immediate local information, thereby compromising measurement accuracy.

\textit{Tactile sensing}, on the other hand, is commonly used to sense contact information focusing on force distribution. Recently, vision-based tactile sensor, such as GelSight~\cite{yuan2017gelsight}, has attracted many people's interest due to their capability to sense fine contact images and estimate F/T distribution, which makes tasks with physical interaction more convenient, such as fabrics or aircraft defect detection\cite{long2021fabric, agarwalrobotic}, object grasping \cite{kanitkar2022poseit, hogan2018tactile}, etc. However, the current vision-based tactile sensing is mainly focused on tabletop perception and manipulation tasks with a quasi-static process.

In this work, we introduce the use of a vision-based tactile sensor on a fully-actuated UAV to enhance aerial manipulation tasks, a setting that is inherently more dynamic than traditional tabletop manipulation, as shown in Fig. \ref{fig: concept}. Specifically, we install a GelSight tactile sensor directly on the end-effector of the aerial manipulator to provide essential force feedback. By fusing with the measurement of the F/T sensor, the contact state estimation gets improved. This enables us to implement a hybrid motion and force control on our aerial manipulator to precisely track reference forces. Additionally, the local perception mechanism of the tactile sensor and its capability to detect surface textures make our system an ideal platform for fine-grained wall texture or defect detection, especially in cases where purely vision-based perception falls short. We also develop the corresponding simulator that integrates vision-based tactile sensor simulation and the fully-actuated UAV simulation in Gazebo\footnote{More details can be found on our project website: \href{https://sites.google.com/view/aerial-system-gelsight}{https://sites.google.com/view/aerial-system-gelsight}}. To the best of our knowledge, we are the first to incorporate a vision-based tactile sensor into aerial manipulation tasks.

In summary, the main contributions of this work are:
\begin{itemize}[leftmargin=*]
	\item We developed a new aerial manipulator system with tactile sensing as the force feedback. To the best of our knowledge, we are the first to incorporate a vision-based tactile sensor into aerial manipulation tasks. 
	
    \item We propose a pipeline that leverages tactile feedback for real-time force tracking using a hybrid motion-force controller on our aerial manipulator. Our approach can replace or supplement traditional force/torque sensors, achieving similar or improved performance.

    \item We propose a method to utilize a vision-based tactile sensor for wall texture detection during aerial interaction. 
    
    \item We present real-world experiments to evaluate the new system in real-time force tracking and wall texture recognition.
\end{itemize}

\section{RELATED WORKS}
\label{sec: related works}

\begin{table}[htbp]
    \centering
    \caption{Comparison between F/T sensor and GelSight tactile sensor}
    \label{tab:sensor_comparison}
    \begin{tabular}{c|c|c}
    \hline
    \hline
\textbf{Characteristic} & \textbf{F/T Sensor} & \textbf{GelSight}\\
\hline
Weight [g]& 255 & 60 \\
Size $L\times W\times H$ [mm] & $75\times 75\times 33$ & $65\times 45\times 45$ \\
Cost [\$] & 5000 & 50 \\
Frequency [Hz] & $>1000$ & $30$ \\
Integration & \makecell{Base \\ (indirect contact)} & \makecell{Tip \\ (direct contact)}\\
\makecell{Contact image} & No & Yes \\
\hline
       \hline 
    \end{tabular}
 \vspace{-1em}
\end{table}

\subsection{Aerial Manipulation}

Aerial manipulation often requires tracking a reference force during contact with external objects to enable complex tasks such as aerial writing~\cite{tzoumanikas2020aerial, lanegger2022aerial} or pushing a target~\cite{brunner2022energy}. 
Prior work includes Bodie et al.\cite{bodie2019omnidirectional,bodie2020active}, who used an F/T sensor and a time-of-flight camera on an omnidirectional aerial manipulator for contact inspection with an axis-selective impedance controller. 
Similarly, Cuniato et al. mounted an F/T sensor on the root of the end-effector and developed a power-based safety layer when pushing a moving cart~\cite{cuniato2022power}. Our previous research~\cite{he2023image} integrated an F/T sensor on a fully actuated UAV and developed hybrid motion and force control with an image-based visual servo for wall painting tasks. Traditional methods for force feedback, such as force/torque or force sensing resistor sensors, have limitations like high cost, bulk, or restricted sensing capabilities. We break new ground by being the first to utilize vision-based tactile sensors on aerial manipulators, offering not only precise force measurements but also texture sensing.

\subsection{Tactile Sensing for Manipulation}

Tactile sensing, typically employed for contact force distribution, comes in various forms, such as tactile arrays and small F/T sensors, as well as multi-modal systems like BioTac~\cite{fishel2012sensing}. Recently, vision-based tactile sensors like GelSight~\cite{yuan2017gelsight} and Digit \cite{lambeta2020digit} have advanced the field, particularly in perception and manipulation tasks. These sensors go beyond force distribution to capture contact images, proving useful in diverse perception applications such as slippage detection \cite{yuan2015measurement}, hardness estimation~\cite{yuan2016estimating}, material identification~\cite{yuan2018active}, and object pose estimation~\cite{villalonga2021tactile}, etc. It also facilitates various manipulation tasks, such as object grasping \cite{kanitkar2022poseit} and re-grasping \cite{hogan2018tactile}, pushing \cite{hogan2020tactile}, cable manipulation \cite{she2021cable}, etc. However, most of the current applications focus on tabletop settings with quasi-static processes. While a few researchers applied tactile sensing to dynamic scenarios, such as liquid property estimation \cite{huang2022understanding}, robot swing \cite{wang2020swingbot}, etc., the application on mobile robots is rare. Our work brings it to aerial manipulation, a more dynamic context.

\section{METHODOLOGY}
\label{sec: controller}

\vspace{-0.1cm}
\begin{figure*}
\centering    
  \includegraphics[width=0.89\textwidth]{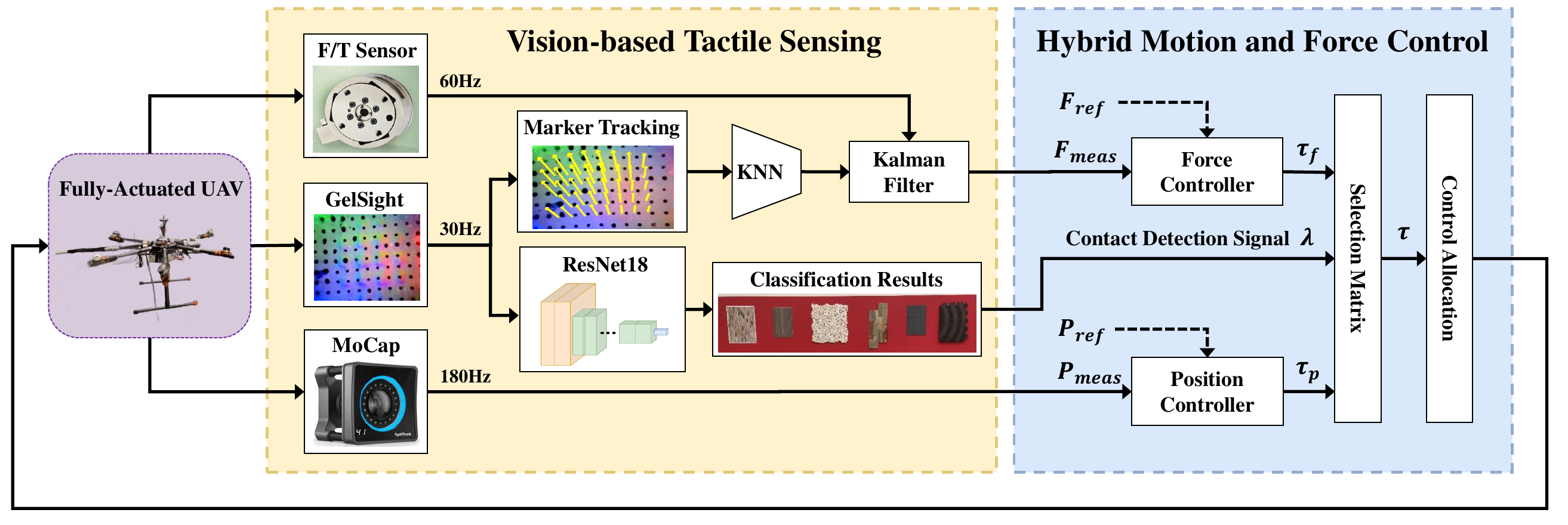}
  \caption{System overview. We developed a pipeline that leverages the vision-based tactile sensor GelSight with the fully actuated UAV. The tactile sensor estimates the contact force and recognizes the wall texture during the flight. The estimated contact force, together with the indirect measurement from the base F/T sensor, is then fused via the Kalman filter to estimate the external contact force for force tracking using a hybrid motion-force controller.}
  \label{fig: system_overview}
\end{figure*}

In this section, we introduce our fully actuated UAV system equipped with a vision-based tactile sensor and the pipeline of tactile sensing-based aerial manipulation, as shown in Fig. \ref{fig: system_overview}.

\subsection{The Design of Aerial Manipulator with Tactile Sensing}

\subsubsection{Aerial Manipulator Design}
To complete the complex aerial manipulation task, we build a fully actuated UAV system based on a hexrotor model (Tarot 960), as shown in Fig. \ref{fig: concept}. Each rotor is tilted at a 30-degree angle, with alternating tilt directions, creating a fully-actuated configuration. We affixed a rigid carbon fiber rod to the front of the vehicle to serve as the manipulator. An ATI Gamma 6-axis F/T sensor is mounted at the base of the manipulator due to its weight (255 g) to balance the system center of mass. The flight control unit is a mRo Pixracer (FMUv4), equipped with a 180 MHz ARM Cortex® M4 processor, an inertial measurement unit (IMU), a barometer, and a gyroscope. Customized PX4 firmware is executed onboard to control the fully actuated platform. Additionally, onboard computation is handled by an Nvidia Jetson TX2, running the ROS Melodic system. For a more comprehensive description of our vehicle's architecture, please refer to our previous work~\cite{he2023image, keipour2022physical}.

\subsubsection{Onboard Tactile Sensor}

In this work, we integrate a vision-based tactile sensor, GelSight, at the tip of the end-effector. GelSight~\cite{yuan2017gelsight} consists of three main components: a dome-shaped soft elastomer gel pad, three RGB LED arrays, and an embedded camera. The external surface of the gel pad is printed with a marker array spaced at 1.7 mm intervals, encompassing approximately 80 markers within the field of view. Upon contact with external objects, the elastomer deforms, reflecting the 3D shape and texture of the contact surface. Concurrently, the printed markers shift, indicating the local force exerted. The RGB LED arrays illuminate the deformed gel pad, and the embedded camera captures images, serving as tactile readings, at a resolution of $320\times240$ pixels resolution and a rate of 30 Hz. In contrast to traditional force/torque (F/T) sensors, GelSight offers advantages in terms of size, weight, cost, and spatial resolution as shown in Table ~\ref{tab:sensor_comparison} and Fig. \ref{fig: concept}. It doesn't require an extra data acquisition device, which most F/T sensor needs. Those features make it convenient to use on the UAV platform.

\subsection{Vision-based Tactile Sensing}

As previously highlighted, the GelSight sensor image contains two key elements: 1) the RGB image, which reveals the 3D texture of the contact surface, and 2) the printed markers on that surface, the motion of which indicates the local force distribution. In our pipeline, we leverage the motion of these markers to estimate the contact force while utilizing the complete RGB image to deduce the surface texture of the contact point.

\subsubsection{Contact Force Estimation}

\cite{yuan2017gelsight} showed that using the CNN model can estimate the 6-axis contact force/torque. However, we note that it is computationally expensive and the robustness of the estimation is not that high during dynamic contact. Instead, we utilize the marker motion to infer the contact force since it represents the deformation of the elastomer, whose relationship to contact force is more straightforward. We first track the motion of specific markers. We capture an initial, non-contact tactile image before takeoff to serve as a reference. Subsequently, we use the optical flow method to track the motion of the central 39 markers at each timestep. This ensures their visibility even under significant shear forces. The two-axis motion of these markers forms a 78-dimensional feature vector. Utilizing the k-nearest neighbors algorithm (kNN), we estimate the contact force at each timestep based on this feature set. Training data, complete with labeled ground-truth forces, are used to facilitate this prediction, as further detailed in Section~\ref{sec:force}. The kNN algorithm identifies the k closest training points to the current marker motion and uses their ground-truth forces to estimate the present contact force.

\subsubsection{Contact Texture Detection}
\label{sec:contact_texture_detection}

We employ the full RGB tactile image from GelSight for contact texture recognition. Specifically, the system determines whether the sensor is in contact with an external surface and, if applicable, identifies its texture. This problem aligns with image classification tasks, treating non-contact as a unique texture category. To address this, we use the ResNet18 structure \cite{He_2016_CVPR} and retrain the network on our dataset. The network takes GelSight images at each timestep as input and outputs the likelihood for each texture category. To enhance prediction robustness, we also incorporate temporal prediction data, assuming that contact textures change infrequently during contact. We define $p_{k,i}$ as the predicted likelihood of the $i^{th}$ texture at timestep $t_k$, and $s_{k, i}$ as the cumulative likelihood for that texture up to $t_k$. The relationship between them is expressed as $s_{k+1,i} = f(s_{k,i},p_{k,i})$, where we empirically define $f(s_{k,i}, p_{k,i}) = 5{s_{k,i}}+{p_{k,i}}$ in this work.

\subsection{Sensor Fusion}

While tactile sensing provides local measurements of force distribution and semantic information, the estimation results from the KNN algorithm may be subject to bias and noise due to sensor wear, aging, and deformation. Conversely, although F/T sensors located at the base of the manipulator measure force indirectly, they offer high-frequency measurements. To leverage the strengths of both sensor types, we employ a standard Kalman filter to fuse the force readings from the tactile sensor $\bm{F}_{tac}$ and the force/torque sensor $\bm{F}_{ft}$ for more accurate force estimation. 

\begin{enumerate}
    \item \textit{State:} The state of the Kalman filter is defined as 
    the contact force $\bm{F}_c$ exerted on the end-effector.
    \item \textit{Process Model:} We assume the UAV is applying a constant push force against the target surface, and express the process model as: \begin{equation}
        \dot{\bm{F}}_c = \bm{0}_3
    \end{equation}
    \item  \textit{Measurement Model:} since the force/torque sensor and the tactile sensor measure the contact force separately, the measurement equation is modeled as follows: \begin{equation}
        \begin{bmatrix}
            \bm{F}_{ft}\\ \bm{F}_{tac}
        \end{bmatrix} = \begin{bmatrix}
            \bm{I}_{3\times 3}\\ \bm{I}_{3\times 3}
        \end{bmatrix} \bm{F}_c + \begin{bmatrix}
            \bm{n}_{ft}\\ \bm{n}_{tac}
        \end{bmatrix}
    \end{equation}
\end{enumerate} where $\bm{n}_{ft}$, $\bm{n}_{tac}\in\mathbb{R}^3$ are the additive Gaussian white noise in the force/torque and tactile sensor measurements.

We follow the established Kalman filter to update the state and covariance estimation. The refined estimations are then fed into the hybrid motion-force controller for the computation of the UAV's subsequent control commands.

\subsection{Hybrid Motion-Force Control Based on Tactile Sensing}

This section introduces the hybrid motion and force control algorithm for the fully-actuated UAV utilizing the estimation of contact force from sensor fusion as the force feedback. We first define the notation and reference frames in Table~\ref{tab:frames}.

{
\renewcommand{\arraystretch}{1.25}
\begin{table}
    \centering
    \caption{Notation Definitions}
    \label{tab:frames}
    \begin{tabular}{c c}
    \hline
    \hline
       \textbf{Symbol}  & \textbf{Definition} \\
       \hline 
        $\mathcal{I}$ &  inertial (world) frame subscript\\
        $\mathcal{B}$ &  body-fixed frame subscript\\
        $\mathcal{E}$ &  end-effector frame subscript\\
        $\mathcal{W}$ &  contact surface (wall) frame \\
        $\bm{R}^A_B$ & rotation in SO(3) from frame B to frame A \\
        \hline
        \hline
    \end{tabular}
    \vspace{-1em}
\end{table}
\renewcommand{\arraystretch}{1}
}
\subsubsection{Aerial Manipulator Dynamics}
The dynamic model of the hexarotor aerial manipulator can be derived following our previous work ~\cite{he2023image}:
\begin{align}
\label{drone-model}
    \bm{M \dot v + C v  + G = \tau' + \tau_{c} }
\end{align}
with inertia matrix $\bm{M}\in\mathbb{R}^{6\times 6}$, centrifugal and Coriolis term $\bm{C}\in\mathbb{R}^{6\times 6}$, gravity wrench $\bm{G}\in\mathbb{R}^{6}$, control wrench $\bm{\tau'}\in\mathbb{R}^{6}$, contact wrench $\bm{\tau}_c \in\mathbb{R}^{6}$ and system twist (linear and angular velocity) $\bm{v} = \begin{bmatrix}
    \Vbf^\top\  & \Omgbf^\top
\end{bmatrix}^\top \in \mathbb{R}^{6}$ expressed in the body frame. 
\begin{align}
    \bm{M} & = \text{diag}\left(\begin{bmatrix}
        m\mathbf{I}_{3\times3} & \bm{J}
    \end{bmatrix}\right)\notag\\
    \bm{C} & = \text{diag}\left(\begin{bmatrix}
        m[\Omgbf]_\times & -\bm{J}[\Omgbf]_\times
    \end{bmatrix}\right)\\
    \bm{G} & = m\text{diag}\left(\begin{bmatrix}
        \mathbf{R}_{\mathcal{I}}^{\mathcal{B}} & \mathbf{0}_{3\times3}
    \end{bmatrix}\right)\bm{g}\notag  
\end{align}
where $m$ is the vehicle mass, $\mathbf{J}$ the moment of inertia, $\mathbf{g}$ the gravity. $[\bm{*}]_\times$ denotes the skew-symmetric matrix associated with vector $\bm{*}$.

Since the system is fully-actuated, we apply the feedback linearization~\cite{shi2019neural}, and the compensated wrench input can be computed as $\bm{\tau}'=\bm{C} v - \bm{G} + \bm{\tau}$. Then, the dynamics of the system \eqref{drone-model} becomes \begin{align}\label{dynamics-linearized}
    &\bm{M}\bm{\dot v} = \bm{\tau} + \bm{\tau}_c
\end{align} 

\subsubsection{Hybrid Motion and Force Control}

Leveraging the advantages of the fully-actuated UAV, the vehicle is capable of controlling its translation and orientation independently, which is different from the coplanar multi-rotors that only generate thrust normal to its rotor plane, leading to completely tilt towards the desired direction. We implement the hybrid motion and force controller in \cite{he2023image}, which combines the output of motion tracking controller $\bm{\tau}_{p}$ and wrench tracking controller $\bm{\tau}_f$ as follows: \begin{align}
    \bm{\tau} &= (\mathbf{I}_{6\times 6}-\bm{\Lambda}) \bm{\tau}_{p} + \bm{\Lambda} \bm{\tau}_f \label{eq:hybrid}\\
    \bm{\Lambda} &= \text{blockdiag}(\bm{\Lambda}', \mathbf{0}_{3\times 3})\in\mathbb{R}^{6\times 6}\\
    \bm{\Lambda}' &= \mathbf{R}_{\mathcal{E}}^\Bframe\begin{bmatrix}
        0 & 0 & 0\\
        0 & 0 & 0\\
        0 & 0 & \lambda\\
    \end{bmatrix}\label{eq:selection}
\end{align} where $\mathbf{R}_{\mathcal{E}}^\Bframe$ denotes the rotation matrix from the end-effector frame to body frame. Intuitively, the matrix $\mathbf{\Lambda}$ selects direct force control commands, and leaves the complementary subspace for the motion control. {$\lambda\in\{0,1\}$ is the force-motion control switch, where $\lambda = 0$ corresponds to the pure motion control, and $\lambda=1$ indicates the force control along the normal direction of the surface. The wrench controller can be activated by tactile contact detection in Sec. \ref{sec:contact_texture_detection}.

To alleviate the oscillation and absorb the pushing energy efficiently, we designed an impedance force control scheme to ensure the end-effector can track reference force $\bm{F}_{ref}$. The interaction force $\bm{F}_{meas}$ acting on the UAV can be measurements from tactile sensors, measurements from force torque sensors, or output of the Kalman filter. The force-tracking control action can be obtained as 
\begin{align}
    \bm{e}_f\! &= \bm{F}_{meas} - \bm{F}_{ref}\notag\\
    \bm{F}_{f}\! &= \! \bm{F}_{ref}\! -\! \bm{M_0}\bm{M}_d^{-1} (\bm{K}_s \bm{e}_s \!+\! \bm{D}_s \dot{\bm{e}}_s)\!+\!\bm{K}_{f,p}\bm{e}_f\! +\! \bm{K}_{f,i}\!\int\!{\bm{e}}_f dt\notag\\
    &\!= \bm{F}_{ref}\! +\! \bm{K}_{s,p} \bm{e}_s\! + \!\bm{K}_{s,d} \dot{\bm{e}}_s \!+ \!\bm{K}_{f,p}\bm{e}_f\! + \!\bm{K}_{f,i}\int\bm{e}_f dt
\end{align} 
with $\bm{e}_s$ the error between current and operating position, $\bm{K}_{f,p}$, $\bm{K}_{f,i}>0$ the positive definite tunable gain matrices, $\bm{M}_0$, $\bm{M}_d$ the actual and desired vehicle mass, respectively. $\bm{K}_{s,p}=\bm{M}_0 \bm{M}_d^{-1} \bm{K}_s$ is the normalized stiffness. $\bm{K}_{s,d}=\bm{M}_0 \bm{M}_d^{-1} \bm{D}_s$ is the normalized damping. Then, the control wrench $\bm{\tau}_f = \begin{bmatrix} \bm{u}_f & \bm{0}_3 \end{bmatrix}^\top\in\mathbb{R}^6$, where $\bm{u}_f = \bm{R}_{\mathcal{W}}^{\Bframe}\bm{F}_{f}$ will be pass through \eqref{eq:selection}.

As for the motion control, the classical geometric multirotor controller is used to obtain the control wrench $\bm{\tau}_p$~\cite{keipour2022physical}. Then, the total control wrench $\bm{\tau}$ is mapped to the rotor speeds of the fully-actuated UAV through the control allocation process.

\section{EXPERIMENTS}
\label{sec: experments}

In this section, we conduct multiple experiments to show the benefit of using the vision-based tactile sensor on the UAV for aerial interaction tasks. Specifically, we perform experiments on force tracking using the hybrid force-motion control, and real-time wall texture recognition on our aerial manipulator leveraging the tactile feedback.

\subsection{Experiment Setup}
We test our developed system and the algorithm pipeline in the real world. An optitrack Motion capture system is used for system localization. For the force tracking and texture recognition experiments, the aerial manipulator is directed to fly against a vertical wall located within the motion capture space, as shown in Fig. \ref{fig: concept}. Each experiment was repeated twice.

\subsection{Contact Force Estimation}\label{sec:force}

We leverage the motion of GelSight markers in each frame to infer contact force. To obtain a more stable dataset, we initially gather data in a controlled tabletop setup and later assess the control performance based on tactile force estimations. In this setup, the GelSight sensor is mounted alongside an F/T sensor, similar to \cite{yuan2017gelsight}. We manually apply varying forces against the sensor using different surfaces. Both the F/T sensor readings and GelSight marker motions are recorded as training data, accumulating a total of 10,000 data points. We also randomly collect 1,300 data points for testing. During experiments, we utilize these data points and employ a kNN method to estimate contact force, where we set $K = 50$. The contact force estimation Root Mean Square Error (RMSE) is around $0.91$ N, which is in a similar error to the work using CNN \cite{yuan2017gelsight}.

\subsection{Hybrid Motion-Force Controller Performance}

\begin{figure}
    \centering
    \includegraphics[width = 0.98\linewidth]{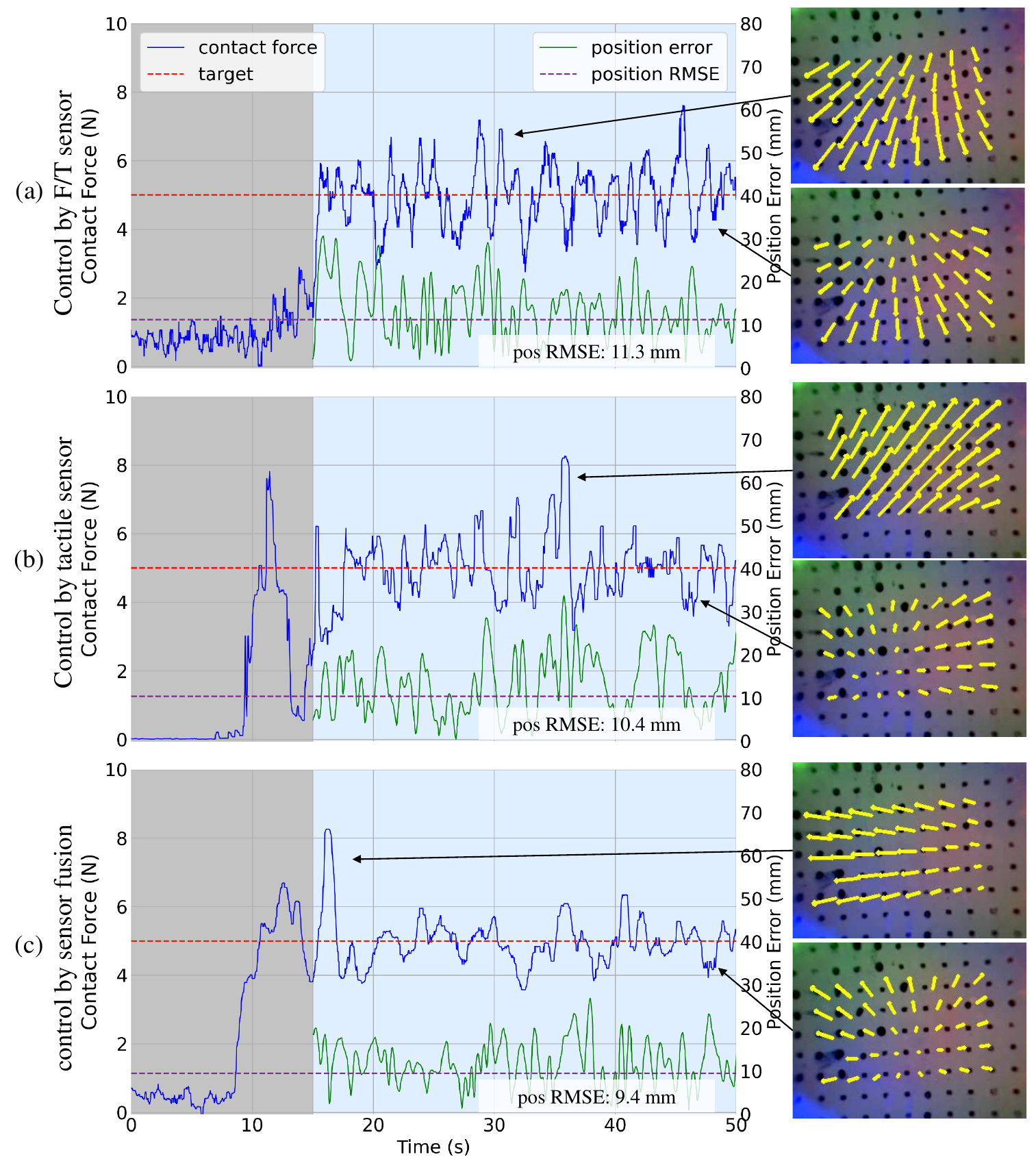}
    \caption{Comparison of controller performance: (a) using F/T sensor only, (b) using vision-based tactile sensor only, and (c) employing sensor fusion. After making contact, the hybrid motion-force controller activates (the blue region) and tracks a constant target contact force.}
    \label{fig: controller_performance}
\end{figure}

In this experiment, we evaluate the performance of our hybrid force-motion controller, which utilizes tactile feedback for force tracking. Upon takeoff, the vehicle is directed to approach a wall and sustain a contact force of $5$ N upon making contact. We compare performance across three scenarios: (1) relying solely on F/T sensor readings, (2) using only vision-based tactile sensor measurements, and (3) employing sensor fusion that combines both types of measurements.

The controller's performance metrics are outlined in Fig. \ref{fig: controller_performance} and Table \ref{tab:controller_performance}. Upon making contact with the wall, the hybrid motion-force controller is engaged. We observed that all three test scenarios achieved the target contact force with near-zero steady-state error. Scenarios relying solely on either the F/T sensor or the vision-based tactile sensor showed similar ranges of error variance. However, the scenario employing fused force state estimates exhibited reduced variance. All scenarios experienced some degree of overshoot and undershoot. A closer examination of the corresponding contact images and 6-axis F/T feedback revealed that large shear forces are present during overshoot periods, suggesting that wrench forces are interrelated during contact. This implies that force/torque control along other axes may require further optimization.

Given the absence of ground truth for contact force (all sensors introduce some level of measurement error), we also use flight performance during force tracking as an indirect yet meaningful metric. As depicted in Fig. \ref{fig: controller_performance} and Table \ref{tab:controller_performance}, we specifically examine position tracking error. Notably, using fused force data leads to about a 16\% improvement in key performance indicators like RMSE, and standard deviation, compared to only relying on the F/T sensor. In summary, the results strongly indicate that the vision-based tactile sensor can effectively serve as a replacement for, or complement to, traditional F/T sensors, yielding comparable or enhanced flight performance.

\begin{table}[ht]
    \centering
    \caption{Controller Performance Comparison}
    \label{tab:controller_performance}
    \begin{tabular}{c|ccc}
    \hline
    \hline
\diagbox{metrics}{controller} & \makecell{Use only \\ F/T sensor} & \makecell{Use only \\ tactile sensor} &\makecell{Use \\ sensor fusion} \\
\hline
Force RMSE (N) & 0.85 & 0.95 & 0.74 \\
Force Overshoot (N) & 2.73 & 3.26 & 2.95 \\
Force Undershoot (N) & 2.76 & 2.69 & 1.49 \\
\hline
Pos RMSE (mm) & 11.28& 10.43 &  \textbf{9.42}\\
Pos Std Dev (mm) & 10.99 & 10.21 & \textbf{8.46}\\
\hline
\hline
        
    \end{tabular}
 \end{table}

\subsection{Wall Texture Recognition}

\begin{figure}
    \centering
    \includegraphics[width = 0.98\linewidth]{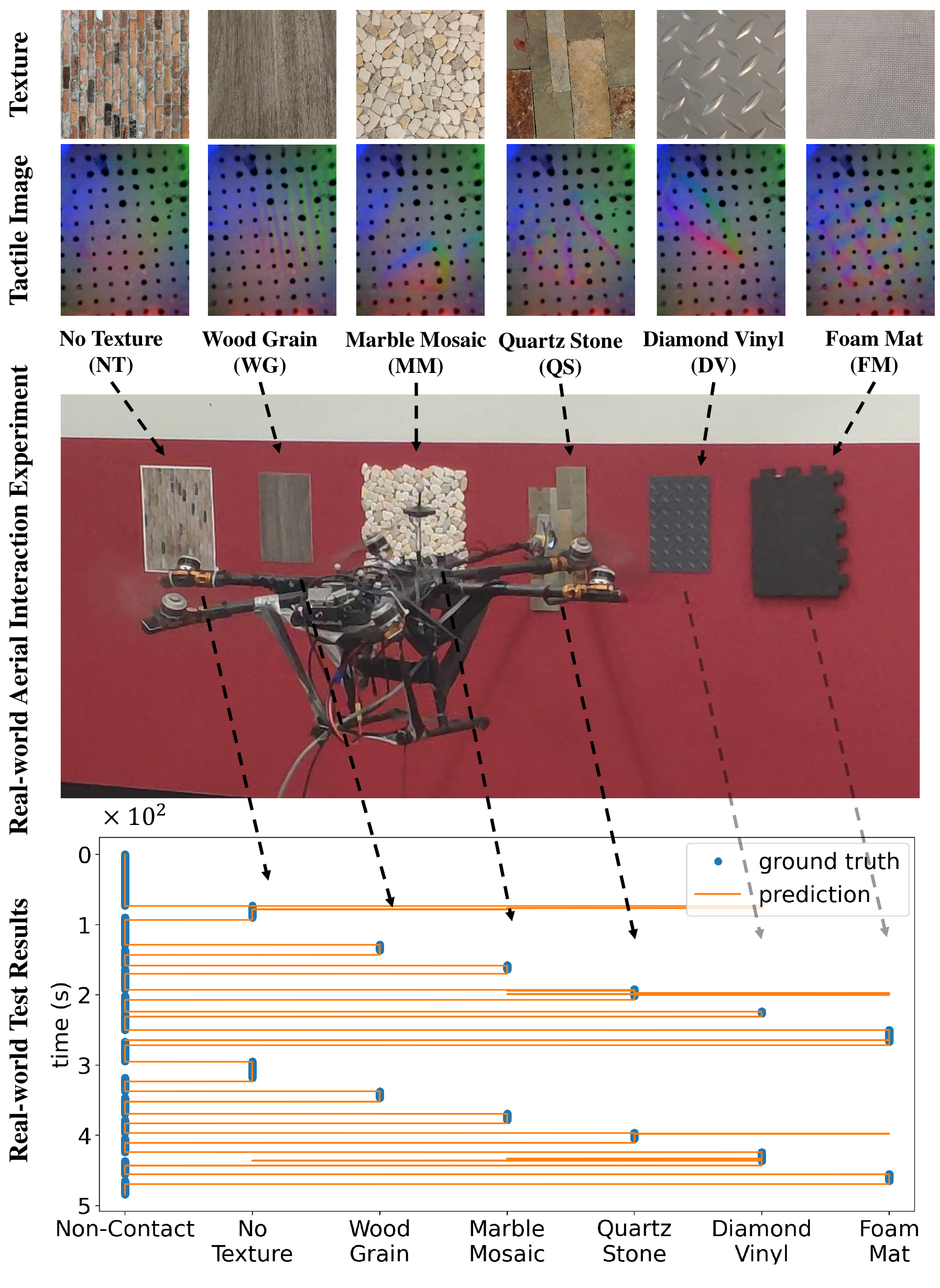}
    \caption{We test our wall texture recognition method on six different textures: (a). No texture, which is a paper with a printed image. (b). Wood grain; (c). marble mosaic; (d) quartz stone; (e) diamond vinyl; (f) foam mat texture. The vehicle makes contact with various textures in a single flight and makes predictions online in real time.}
    \label{fig: wall_texture_detection}
\end{figure}

In this experiment, we evaluate the system's ability to identify wall textures during flight and contact. As shown in Fig. \ref{fig: wall_texture_detection}, we focus on identifying five specific textures: 1) wood grain, 2) marble mosaic, 3) quartz stone, 4) diamond vinyl, and 5) foam mat. Additionally, we include a sixth texture: a printed imitation of a brick pattern on flat paper. This mimics complex 3D textures while actually being a flat surface, akin to many wallpapers, highlighting the need for tactile feedback beyond simple visual inspection for accurate texture and material perception. We also consider the absence of contact as an additional texture category. To build a robust training dataset, we manually collected samples by pressing the GelSight tactile sensor against these various textures. Each texture contributed 3,000 data points to the training set, which includes all frames where contact was made.

During the flight tests, the aerial manipulator is flown towards various textures affixed to a wall, making contact with approximately 5 N of force for at least 10 seconds. This duration allows the trained classifier to identify and categorize the texture in contact. The manipulator then proceeds to the next target texture. Each texture is engaged twice in this manner. The entire procedure is executed in a continuous flight, without landing the UAV in between tests.

\begin{figure}
    \centering
    \includegraphics[width = 0.98\linewidth]{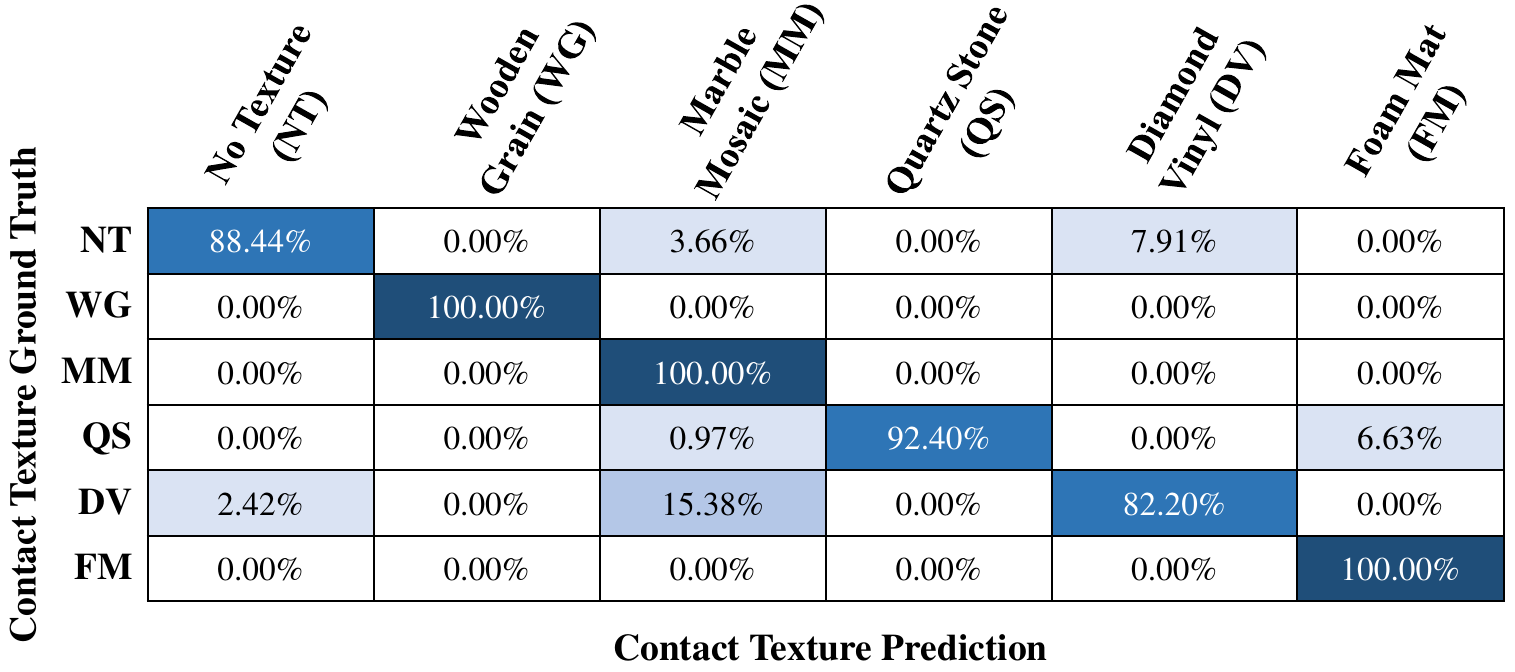}
    \caption{Confusion matrix for real-world contact texture recognition at each timestep during flight tests. The average classification accuracy is 93.4\%. }
    \label{fig:texture_classification}
\end{figure}

The prediction result is shown in Fig. \ref{fig: wall_texture_detection} and Fig. \ref{fig:texture_classification}. We note that most of the time, the prediction is accurate with an average accuracy of 93.4\%. There are brief inconsistencies in predicting certain textures such as the diamond vinyl, where the sensor momentarily confuses them with marble mosaic texture. These deviations are understandable given the textural similarities with close features. Importantly, the predictions stabilize and self-correct over time, ultimately achieving a post-contact accuracy of 100\% for all textures. These results demonstrate that the vision-based tactile sensor can effectively identify 3D textures during aerial contact, suggesting its potential utility in defect detection tasks, such as identifying wing cracks, as depicted in Fig. \ref{fig: concept}.
\section{CONCLUSION}
\label{sec: conclusion}

This paper develops a fully-actuated aerial platform equipped with a vision-based tactile sensor for aerial manipulation tasks. We design a pipeline that integrates tactile feedback into a hybrid motion-force controller for real-time force tracking during aerial physical interaction. Additionally, we developed an algorithm for wall texture identification based on tactile measurement. Real-world experimental results indicate that our system can either supplement or replace traditional force/torque sensors while maintaining or improving control performance. Utilizing a fused force-state estimate, we achieve a 16\% improvement in position tracking error compared to single-sensor configurations. Furthermore, our tactile sensing method demonstrates a 93.4\% accuracy rate in real-time texture classification, rising to 100\% post-contact. These results highlight the value of incorporating vision-based tactile sensors in aerial manipulation tasks. Future directions include optimizing sensor design and devising a controller that leverages high-dimensional tactile data instead of low-dimensional estimated force/torque values.

\bibliographystyle{IEEEtran}
\bibliography{root}

\end{document}